%% file: main.tex
\documentclass[10pt,twocolumn,letterpaper]{article}

\usepackage{cvpr}              


\definecolor{cvprblue}{rgb}{0.21,0.49,0.74}
\usepackage[pagebackref,breaklinks,colorlinks,allcolors=cvprblue]{hyperref}
\usepackage{tikz}
\usepackage{multirow}
\usepackage{graphicx}
\usepackage{float}
\def\paperID{1540} 
\def\confName{CVPR}
\def\confYear{2025}

\newcommand{\Ours}{Drive-1-to-3 }

\title{Drive-1-to-3: Enriching Diffusion Priors for Novel View Synthesis \\ of Real Vehicles}

\author{Chuang Lin$^1$ \quad Bingbing Zhuang$^4$ \quad Shanlin Sun$^2$ \quad Ziyu Jiang$^4$ \\
Jianfei Cai$^1$ \quad Manmohan Chandraker$^{3,4}$  \vspace{+0.3em} \\
$^1$Monash University~~~$^2$UC Irvine~~~$^3$UC San Diego~~~$^4$NEC Labs America\vspace{-0em} \\
}

\begin{document}

\twocolumn[{%
\renewcommand\twocolumn[1][]{#1}%
\maketitle
\vspace{-8mm}
\begin{center}
    \centering
  \includegraphics[width=0.9\linewidth]{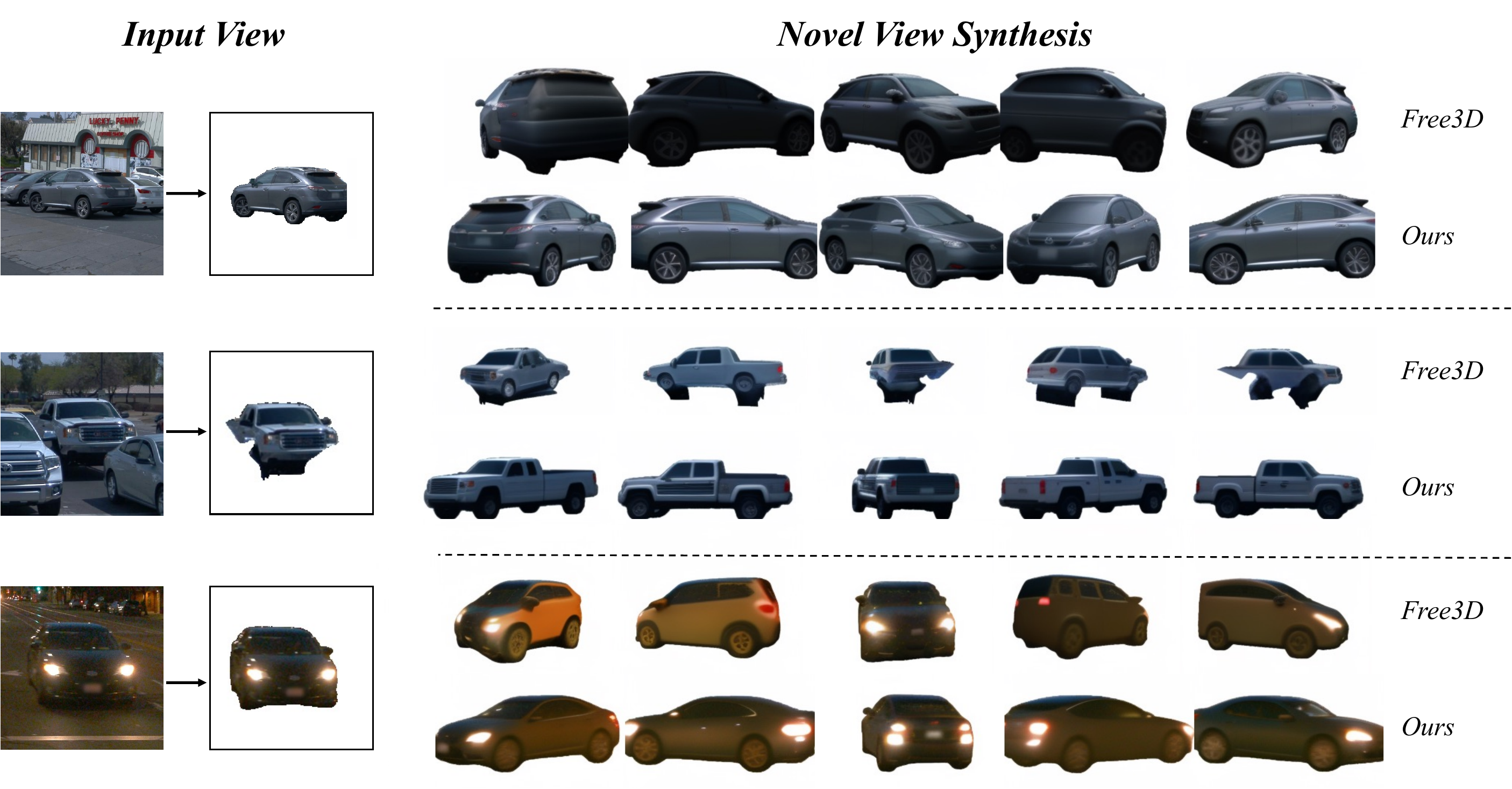}
  \captionof{figure}{Comparisons between the pretrained Free3D~\cite{zheng2023free3d} and ours on real vehicle images, demonstrating our large performance gain.} 
  \label{fig:teaser}   
\end{center}%
}]

\input{sec/0_abstract}    
\input{sec/1_intro}

\input{sec/2_relatedwork}
\input{sec/3_method}
\input{sec/4_experiments}

\input{sec/5_conclusion}

{
    \small
    \bibliographystyle{ieeenat_fullname}
    \bibliography{main}
}

\input{sec/X_suppl}

\end{document}

%% file: sec/0_abstract.tex
\begin{abstract}
The recent advent of large-scale 3D data, e.g. Objaverse~\cite{deitke2023objaverse}, has led to impressive progress in training pose-conditioned diffusion models for novel view synthesis. However, due to the synthetic nature of such 3D data, their performance drops significantly when applied to real-world images. This paper consolidates a set of good practices to finetune large pretrained models for a real-world task -- harvesting vehicle assets for autonomous driving applications. To this end, we delve into the discrepancies between the synthetic data and real driving data, then develop several strategies to account for them properly. Specifically, we start with a virtual camera rotation of real images to ensure geometric alignment with synthetic data and consistency with the pose manifold defined by pretrained models. We also identify important design choices in object-centric data curation to account for varying object distances in real driving scenes -- learn across varying object scales with fixed camera focal length. Further, we perform occlusion-aware training in latent spaces to account for ubiquitous occlusions in real data, and handle large viewpoint changes by leveraging a symmetric prior. 
Our insights lead to effective finetuning that results in a $68.8\%$ reduction in FID for novel view synthesis over prior arts.
\end{abstract}

%% file: sec/1_intro.tex
\section{Introduction}
\label{sec:intro}

Applications like autonomous driving require extensive training and validation in real-world scenarios, which limits scalability. Simulation \cite{yang2023unisim,tonderski2023neurad,guo2023streetsurf,wu2023mars} offers an alternative to large-scale data curation and road testing. Given that traditional simulation pipelines require costly manual work from graphics artists, there is a strong demand for automatically harvesting digital 3D object assets from real driving logs. Novel view synthesis (NVS) emerges as an important cornerstone towards this goal, as it enables the generation of unseen perspectives of objects from limited input views. This capability is essential for ultimately achieving high-quality 3D reconstructions.

Object-level novel view synthesis with pose conditioning has made striking progress in recent years. This is largely attributed to Objaverse~\citep{deitke2023objaverse} that presented an unprecedented million-scale dataset of Internet-sourced 3D models. Such data has inspired the pioneering work of Zero-1-to-3 \citep{liu2023zero} and its follow-ups (\eg ~\cite{zheng2023free3d,liu2023syncdreamer,shi2023zero123++})  to inject 3D knowledge in the form of camera pose conditioning into an image generative model like Stable Diffusion~\citep{rombach2022high}. Such a model is adept at learning generic 3D priors due to the usage of diverse 3D data. But the synthetic nature of Objaverse causes the model to stumble when applied to real-world data, resulting in cartoonish images with poor pose alignment, as shown in \cref{fig:teaser}. 
Our goal in this work is to effectively transfer such 3D priors to a domain-specific application focused on novel view synthesis of real vehicles.

The nature of driving data gives rise to serval challenges -- 1) the lack of ground truth 3D model; 2) multiple views along a temporal trajectory have limited viewpoint changes due to the largely restricted forward motions in driving; 3) heavy occlusions especially under crowded traffic scenes. All these lead to a high degree of ambiguity, and despite existing works \citep{xu2023discoscene,muller2022autorf,shen2023gina} learning categorical prior across instances to tackle it, performances have remained limited in image quality despite test-time optimization. Our key contribution is a consolidated set of good practices to effectively finetune diffusion models on vehicles in real-world road scenes, thereby tackling the aforementioned challenges with the support of rich 3D priors.

Specifically, we start by 
noting the simplified imaging configuration in synthetic data falls short in modeling real driving data. Our first step to close the gap is by projecting real camera poses into object-centric orbital camera poses,
which is of significance as it allows to inherit the rich generic prior in pretrained models in a geometrically consistent manner. Next, to handle large variations in camera-to-object distances in real data, 
we identify constant focal length as an important design choice when cropping object-centric image patch from the original images, as varying object scales are more feasible for the network to tackle  compared to varying focal lengths.
Further, 
our framework accounts for the inevitable presence of occlusions in the source view, by converting the occlusion mask to the latent space where the denoising process operates, and performing occlusion-aware loss supervision to disregard occluded regions on the image. Finally, driving sequences give rise to limited viewpoint change in training data, hence hindering the performance when the target view deviates significantly from the source view. To address this, we enlarge the pose variations in training data by leveraging a left-right symmetric prior for vehicle categories, which yields strong benefits for view synthesis under large viewpoint changes. 
Our insights lead to an effective domain-specific finetuning framework, Drive-1-to-3, on top of the pretrained model~\cite{zheng2023free3d}, yielding significant performance gain as shown in \cref{fig:teaser}.

\noindent As a summary of contributions, we delve into the discrepancy between the Objaverse data and real driving data from various aspects, thereby designing a geometry-aware fine-tuning pipeline at a low training expense, which outperforms prior arts in NVS by a large margin. We further demonstrate in \cref{sec:exp} the application of Drive-1-to-3 in downstream 3D reconstruction and virtual object insertion.

%% file: sec/2_relatedwork.tex
\section{Related Work}
\begin{figure*}[t!]
  \centering
  \vspace{-0cm}
  \includegraphics[width=1.0\linewidth, trim = 0mm 0mm 0mm 0mm, clip]{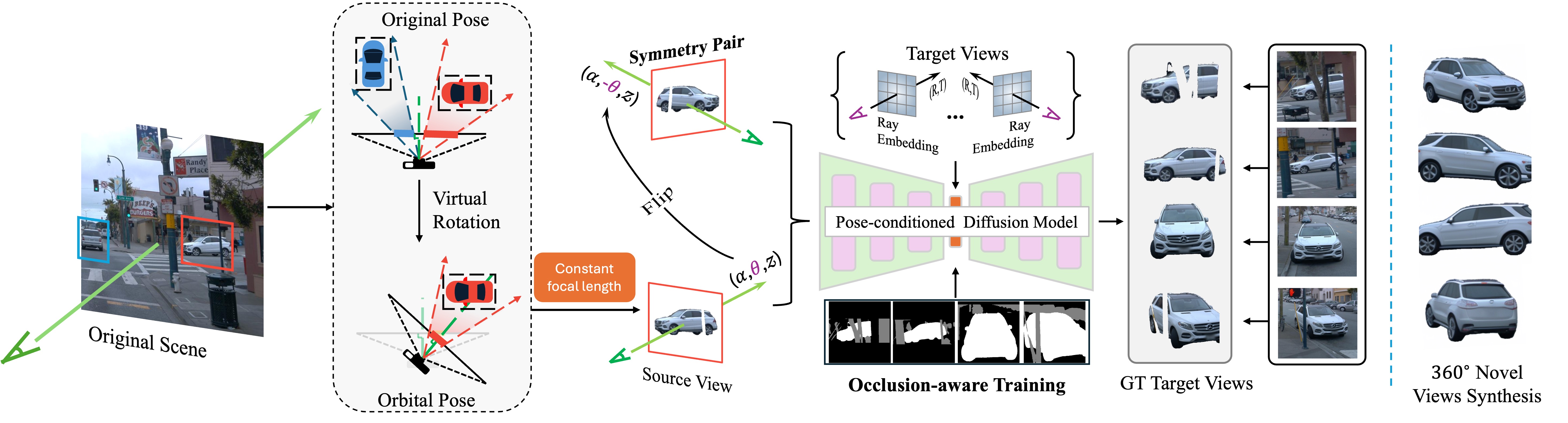}
  \centering
  \caption{\textbf{The overall pipeline of Drive-1-to-3.} First, it processes a single vehicle image from on-board cameras, virtually rotating it to a shared orbital pose. The object-centric image cropped with a constant focal length is fed to a pose-conditioned diffusion model, which performs occlusion-aware training in latent space with a symmetric prior.  }
  \vspace{-2mm}
  \label{fig:overview}   
\end{figure*}

\textbf{Pose-conditioned diffusion models.~} The advent of large-scale 3D data Objaverse~\cite{deitke2023objaverse}  has stimulated many explorations on injecting 3D knowledge into large-scale pretrained image generative models, such as Stable Diffusion~\cite{rombach2022high}. Zero-1-to-3~\cite{liu2023zero} and 3DiM~\cite{watson2022novel} represent the pioneering work along this line, which transform diffusion models into a novel view synthesis machine conditioned on an input image and the camera pose of the target view. This attains impressive view synthesis quality from just one single image as it inherits the rich and generic image prior in the 2D diffusion model that is learned from billions of images. Subsequent works have improved Zero-1-to-3 in various aspects. Free3D~\cite{zheng2023free3d} and EscherNet~\cite{kong2024eschernet} increases the pose conditioning accuracy; Syncdreamer~\cite{liu2023syncdreamer}, Consistent123~\cite{weng2023consistent123}, Wonder3d~\cite{long2023wonder3d} and  others~\cite{shi2023mvdream,tang2024mvdiffusion++,woo2023harmonyview} focus on multiview consistency. One-2-3-45~\cite{liu2024one,liu2023one} improves the efficiency. 
More recent works such as SV3D \cite{voleti2025sv3d} and IM-3D~\cite{melas20243d} adopt video diffusion models for this task thanks to its inherent smoothness across frames.  
However, the above works primarily rely on the synthetic Objaverse~\cite{deitke2023objaverse} data for training, leading to poor performance when applied to real data. 
In this work, we build on top of Free3D~\cite{zheng2023free3d} and research good practices to translate the rich diffusion prior learned from large-scale synthetic data further to imperfect real-world vehicle images captured in the wild.

\noindent \textbf{Large 3D reconstruction models.~} 
Another line of research~\cite{zou2024triplane,han2025vfusion3d,zhang2025gs,wang2025crm} leverages supervision from  3D data~\cite{deitke2023objaverse} and directly predicts a 3D representation in a single forward pass.
LRM~\cite{hong2023lrm} adopts a transformer-based architecture that converts a single image into a NeRF; it is further extended to take multiple views as input~\cite{xu2023dmv3d}, unposed images~\cite{wang2023pf}, and text-to-3D~\cite{li2023instant3d}. Similarly, GRM~\cite{xu2024grm} and LGM~\cite{tang2024lgm} recover a Gaussian Splatting model from sparse-view images. A concurrent work Real3D~\cite{jiang2024real3d} proposes large reconstruction models for real-world images, but not handling vehicles in real driving scene. By combination with LGM, we demonstrate in this work that Drive-1-to-3 can benefit the downstream 3D reconstruction task as a multiview view synthesis machine.

\noindent \textbf{3D learning for real-world vehicles.} 
Single-image based 3D reconstruction of real-world vehicles has remained an open problem due to its inherent ambiguity, the lack of high quality 3D models as ground truth, and imperfect data capture such as occlusions. Inspired by the success of pixelNeRF~\cite{yu2021pixelnerf}, recent works leverage encoder-decoder based networks to learn categorical priors, and regress a NeRF representation from a single vehicle image; examples are AutoRF~\cite{muller2022autorf}, CodeNeRF~\cite{jang2021codenerf}, and Car-Studio~\cite{liu2023car}.
Some research~\cite{shen2023gina,xu2023discoscene,violante2024physically} also explores enhancing features with 3D GANs and tri-plane representations, as well as with more fine-grained material properties. However, the performance remains as yet limited due to its ambiguious nature.
In this work, we present the first attempt of its kind to leverage the rich priors in pre-trained diffusion networks to boost the performance in view synthesis. We also notice a concurrent work Neural Assets~\cite{wu2024neural} capable of view synthesis for real vehicles together with background, but our work distinguishes itself by focusing on finetuning large pretrained models for real data that obviates the need for extensive computing resources. We also note a concurrent effort~\cite{du20243drealcar} to collect a dataset for real vehicles with surrounding views; the dataset may well benefit our task but is pending release.

%% file: sec/3_method.tex
\section{Method}

\subsection{Overview}
\label{sec:overview}
\noindent \textbf{Pipeline.} An overview of our \Ours is illustrated in \cref{fig:overview}. It takes as input a single image patch of a vehicle cropped from the full image captured by onboard cameras. We first apply a virtual camera rotation (\cref{sec:method_pose}) in order for the image crop to have a orbital camera pose. We then flip the image left-right to compose a symmetric pair (\cref{sec:symmetry}) as training input. We account for occlusions (\cref{sec:occlusion}) in the latent space of the pose-conditioned diffusion model.
We use other available views of the object in the same driving sequence as ground-truth target view for network training. 
During inference, the diffusion model may be conditioned with pose to generate $360^\circ$ novel views of the object.

\noindent \textbf{Backbone.} We take Free3D~\cite{zheng2023free3d}, which is finetuned upon Zero-1-to-3~\cite{liu2023zero}, as our diffusion backbone. 
Zero-1-to-3 conditions the diffusion model on the relative pose between the source and targe view, denoted as $[\textbf{R}, \textbf{T}]$ for rotation and translation. Besides this global pose condition, Free3D~\cite{zheng2023free3d} adds per-pixel local pose condition, represented as a map of Pl\"ucker ray embedding $\textbf{f}_i$ for each pixel $i$: 
\vspace{-0.1cm}
\begin{equation}
    \textbf{f}_i = \{\textbf{o}_i\times \textbf{d}_i,\textbf{d}_i\},
\end{equation}
where $\textbf{o}_i$ and $\textbf{d}_i$ denote the ray origin and direction relative to the source view. The ray embedding maps are applied to modulate the image latents in U-Net for improving pose conditioning. We refer readers to \cite{zheng2023free3d} for more details.

\subsection{Canonical Pose Space}
\label{sec:method_pose}

In this section, we explain how the simplified imaging configuration in synthetic data fails to accurately model real data and propose our solutions to address this issue.

\vspace{0.5mm}
\noindent \textbf{Camera pose characteristics in pretrained large models.~} In order for a canonical object-centric view that facilitates learning, it has been a common practice to render training images from Objaverse~\cite{deitke2023objaverse} 3D models in \textit{orbital camera poses} -- camera optical axis points to the origin in world coordinate (i.e. object center) with upright camera orientation along z-axis (i.e. the gravity direction). As such, the camera pose $\mathbf{\Pi}$ w.r.t. the object is uniquely defined by the reduced three degrees of freedom - elevation $\alpha$, azimuth $\theta$, and distance $z$:
\vspace{-0.1cm}
\begin{equation}
    \mathbf{\Pi}=(\alpha,\theta,z).
\end{equation}
They are further used in Zero-1-2-3~\cite{liu2023zero} to describe the relative pose between the source and target view as pose condition for the diffusion model. Besides viewpoint orientation, the object is always placed in a near field with small distance variations in front of the camera, having a large extent of the rendered object with high resolution. Finally, images are rendered with same camera intrinsics, manifested as a fixed field of view.

\begin{figure}[t]
  \centering
  \vspace{-0.5cm}
  \includegraphics[width=1.0\linewidth, trim = 0mm 0mm 0mm 0mm, clip]{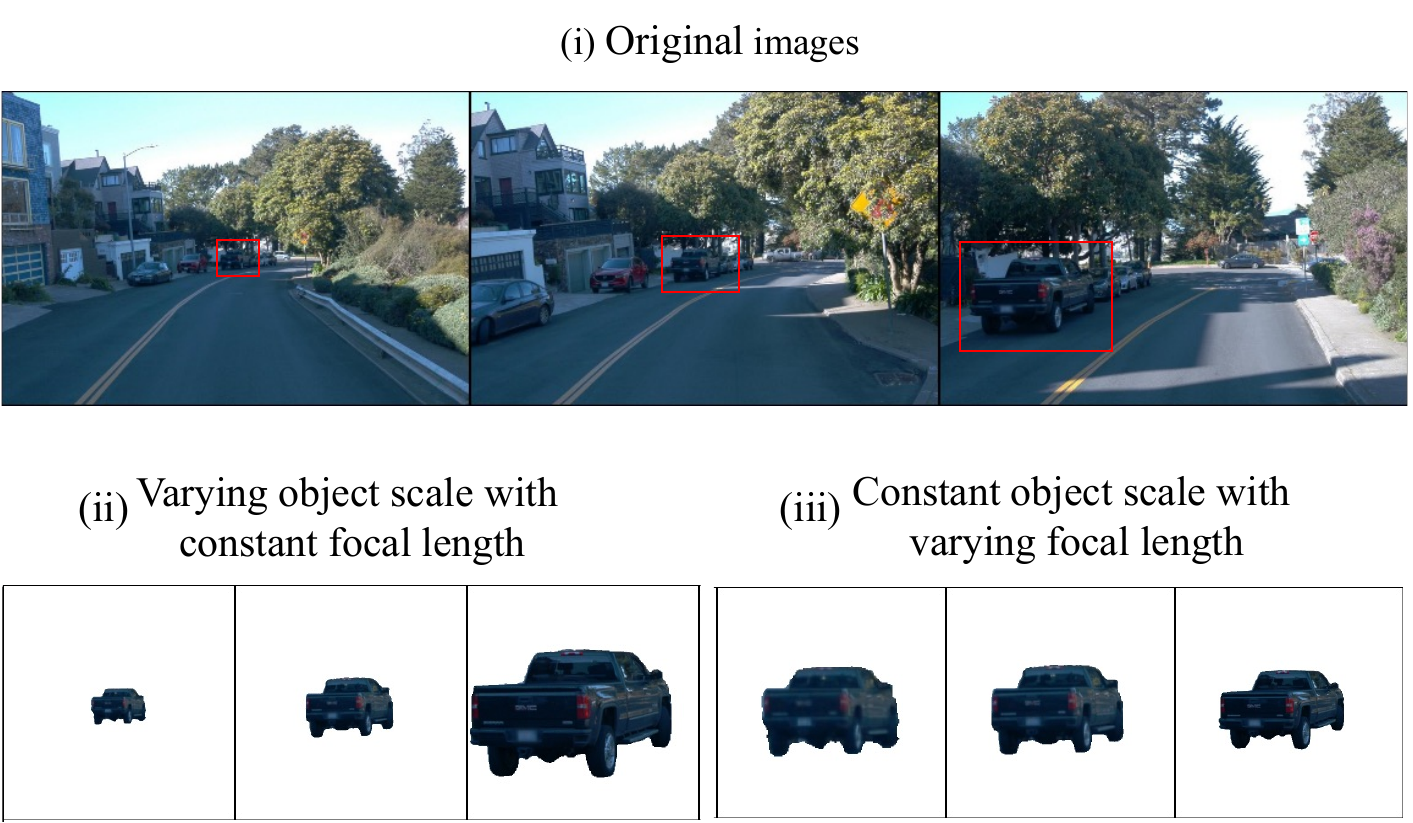}
  \vspace{-0.6cm}  
  \caption{Illustration of two strategies in object-centric image cropping -- varying object scales vs. varying focal lengths.}
  \label{fig:virtualrotandfov}   
\end{figure}

\vspace{0.5mm}
\noindent \textbf{Camera pose characteristics in real data.~} The camera pose distribution in real data deviates largely from the canonical pose space in Objverse data. First, on-board cameras capture multiple objects in the scene simultaneously, without the optical axis passing though object centers as in orbital camera poses, as illustrated in \cref{fig:overview}. Second, the object-camera distances vary significantly from the near to far field, e.g. from 2 to 100 meters in real traffic scene; this results in large variations in the objects' extent in the image plane, as illustrated in \cref{fig:virtualrotandfov} (i). Next, we discuss our handling of these two sources of gaps.

\vspace{1.5mm}
\noindent \textbf{Object-centric virtual camera rotation.} In order to inherit the strong pose conditioning prior from the pretrained large models in a geometry-aware manner, we first transform the camera pose into an orbital one as a canonical pose space. As illustrated in \cref{fig:overview}, for each object in the scene, we virtually rotate the camera to be congruent with an orbital camera pose. With the camera center unaltered, this transformation is scene-independent and pixels can be warped precisely with a rotational homography~\citep{hartley2003multiple}, which is a widely used transformation in geometric vision such as stereo rectification~\citep{loop1999computing}. Consequently, this step creates object-centric images as in Objaverse data, and remarkably, allows to depict the camera pose in the format of $(\alpha,\theta,z)$.

\vspace{1.5mm}
\noindent \textbf{Varying object scales vs. varying focal lengths.} Next, we explore two strategies in choosing the field of view to crop the object patch as illustrated in \cref{fig:virtualrotandfov}, with a view to handling the varying object scales in real images. 

\noindent 1) \textit{varying object scales} -- we use a fixed field of view as in Objaverse data, leaving the object scale variation as is in the cropped object patch. 

\noindent 2) \textit{varying focal lengths} -- we determine the field of view adaptively so as to have similar object scales across all images; we achieve so via expanding the ground-truth 2D bounding box of the object by a fixed ratio followed by a squared cropping and resizing. With a fixed image size ($256 \times 256$), the varying field of view effectively translates to varying focal lengths in the resultant images. 

Empirically, we find that the first strategy with fixed focal length and varying object scales yields far more stable results. This may be explained by the image encoder pretrained on billions of images being inherently adept at learning robust representations across different scales. 
Conversely, the concept of varying focal lengths and their resulting perspective effects are not learned during pretraining on the Objaverse data, posing challenges when learning exclusively from real data.
Specifically, as we shall detail in \cref{sec:exp}, the model trained in this way leads to noticeable object size inconsistency in the generated images. 

\vspace{1.5mm}
\noindent \textbf{Relative pose vs. absolute pose.} As an Internet-sourced dataset, the 3D models in Objaverse are aligned only in elevation (approximately along gravity direction), but not in azimuth, which is randomly specified by model creators without a common reference. Hence, as noted in \citep{shi2023zero123++,liu2024one}, Zero-1-to-3 and its successors can only rely on the relative pose with respect to the source view as the condition, rather than the more informative absolute poses that independently characterize both the source and target view poses. We illustrate this distinction with more detailed explanations in the supplementary material. Unlike Objaverse,  real-world autonomous driving datasets, such as Waymo \citep{sun2020scalability}, do have human annotated object 3D boxes, yielding aligned absolute poses. However, adopting absolute pose condition introduces a trade-off -- 
it forgoes the generic pose-conditioning prior learned from Objaverse's relative poses and instead must learn directly from scratch using only real data.
In this paper, we empirically examine this trade-off by applying absolute poses in the pose-conditioning layers, while retaining other pretrained network parameters. Overall, we observe that the rich priors in the pretrained model outweigh the benefits brought about by the absolute poses.

\subsection{Occlusion Handling}
\label{sec:occlusion}
Here, we discuss our handling of occlusions, a unique challenge for real data captured in the wild.

\noindent \textbf{Occlusion masks.~} In the absence of ground truth occlusion masks, we rely on semantic segmentation to identify plausible occluding regions -- neighboring movable objects and tress/poles are likely occluding the object of interest, whereas sky, road surface, and buildings may be safely deemed as background, as illustrated with an example in \cref{fig:overview}; we leave more details to the supplementary and also refer to AutoRF~\citep{muller2022autorf} for similar operations. 

\vspace{1.5mm}
\noindent \textbf{Occlusion-aware training.~} Our goal then boils down to preventing the potential occlusion from impacting the view synthesis. A natural way to do so is by removing such pixels in the training loss. But the use of a latent diffusion model complicates the problem, as the denoising and hence the noise prediction loss are both applied in the \textit{latent space} instead of on the pixels themselves - strictly speaking, no exact one-to-one correspondences exist for mapping pixels to the elements in the latent feature map due to the receptive field of networks. However, recent studies~\citep{avrahami2023blended} have shown evidence that the masking operation in latent space transfers seamlessly to image space for the image inpanting task. We do the same in our task. 
Specifically, denoting the encoded source and target view in latent space as  $z^{\text{src}}$ and $z^{\text{trg}}$, we first downsample the target view occlusion mask $m$ to the same size as $z^{\text{trg}}$, denoted as $m_d$. The standard loss between predicted noise $\epsilon_\theta$ and its ground truth $\epsilon$  in Zero-1-to-3 is updated to
\begin{equation}
    \mathcal{L} = \mathbb{E}_{(z^{\text{trg}}, z^{\text{src}}, \mathbf{P}), \epsilon, t} \left[ \| \text{M}(\epsilon, m_d) - \text{M}(\epsilon_\theta(z^{\text{trg}}, t, y), m_d) \|_2^2 \right],
\end{equation}
where the network is conditioned on $y = (z^{\text{src}}, \mathbf{P})$, with $\mathbf{P}$ including both the global and local pose embedding, and $\text{M}(x, m_d) = x \cdot m_d + (1 - m_d)$ is used to apply the occlusion mask. We empirically observe this strategy to be fairly effective, as shown in \cref{sec:exp}. Besides the loss, we also attempted to concatenate the source image with its occlusion mask in the network input, with a view to supplying direct occlusion signal. We did not observe noticeable improvement, plausibly due to the powerful pretrained U-Net already being capable of learning occlusion reasoning from the input image itself.

\subsection{Large Viewpoint Changes}
\label{sec:symmetry}

Synthetic data like Objaverse~\cite{deitke2023objaverse} possesses full 3D models that can be rendered from any arbitrary views, enabling large viewpoint changes between the source and target view in both training and inference. 
In contrast, real vehicles captured in driving scenes often exhibit limited viewpoint variation throughout the sequence, resulting in a scarcity of training pairs with large viewpoint changes, which further impacts inference.
Formally, denoting the pose in the source and target view as $\mathbf{\Pi}^{src}=(\alpha
^{src},\theta^{src},z^{src})$ and  $\mathbf{\Pi}^{trg}=(\alpha^{trg},\theta^{trg},z^{trg})$, a large viewpoint change refers to primarily the discrepancy in azimuth $|\theta^{src}-\theta^{trg}|$.
To mitigate this issue, we leverage the symmetric nature of the vehicle category during training. As illustrated in \cref{fig:overview}, we fix the target view, whereas obtaining the symmetric counterpart for the object instance in the source view by horizontally flipping the image and change the camera pose from $(\alpha^{src},\theta^{src},z^{src})$ to $(\alpha^{src},-\theta^{src},z^{src})$, which effectively increases the viewpoint change from the target view, 

As one may notice, this strategy is essentially promoting symmetry in the underlying object representation. While the network might learn the symmetric prior from data by itself, we find it beneficial to explicitly enforce this prior during training in order to achieve pose consistency under large viewpoint changes, and the strength of this guidance makes a difference. Specifically, we have attempted two strategies with varying degree of guidance:

\noindent 1) \textit{weak guidance} as a standard data augmentation, where  each image and its camera pose are horizontally flipped with a $50\%$ probability before being fed into the network. 

\noindent 2) \textit{strong guidance} by training the network with pairs of symmetric images, 
where each image instance and its symmetric counterpart are placed in the same batch, providing a stronger foundation for prompting gradients that account for both small and large viewpoint changes.

\noindent Empirically, we observe that strong guidance leads to significantly superior capability in handling large viewpoint changes, underscoring the effectiveness of such guidance in expanding the span of pose variations during training.

\paragraph{Compute Efficiency.} 
Our domain-specific finetuning achieves large gains at a lower training expense compared to training from scratch for the new domain.
It takes only ten hours with a single A6000 GPU, in contrast to Zero-1-to-3 training for seven days on 8 A100 GPUs. 

%% file: sec/4_experiments.tex
\section{Experiments}
\label{sec:exp}

\begin{figure}[t]
  \centering
  \includegraphics[width=1.0\linewidth, trim = 0mm 0mm 0mm 0mm, clip]{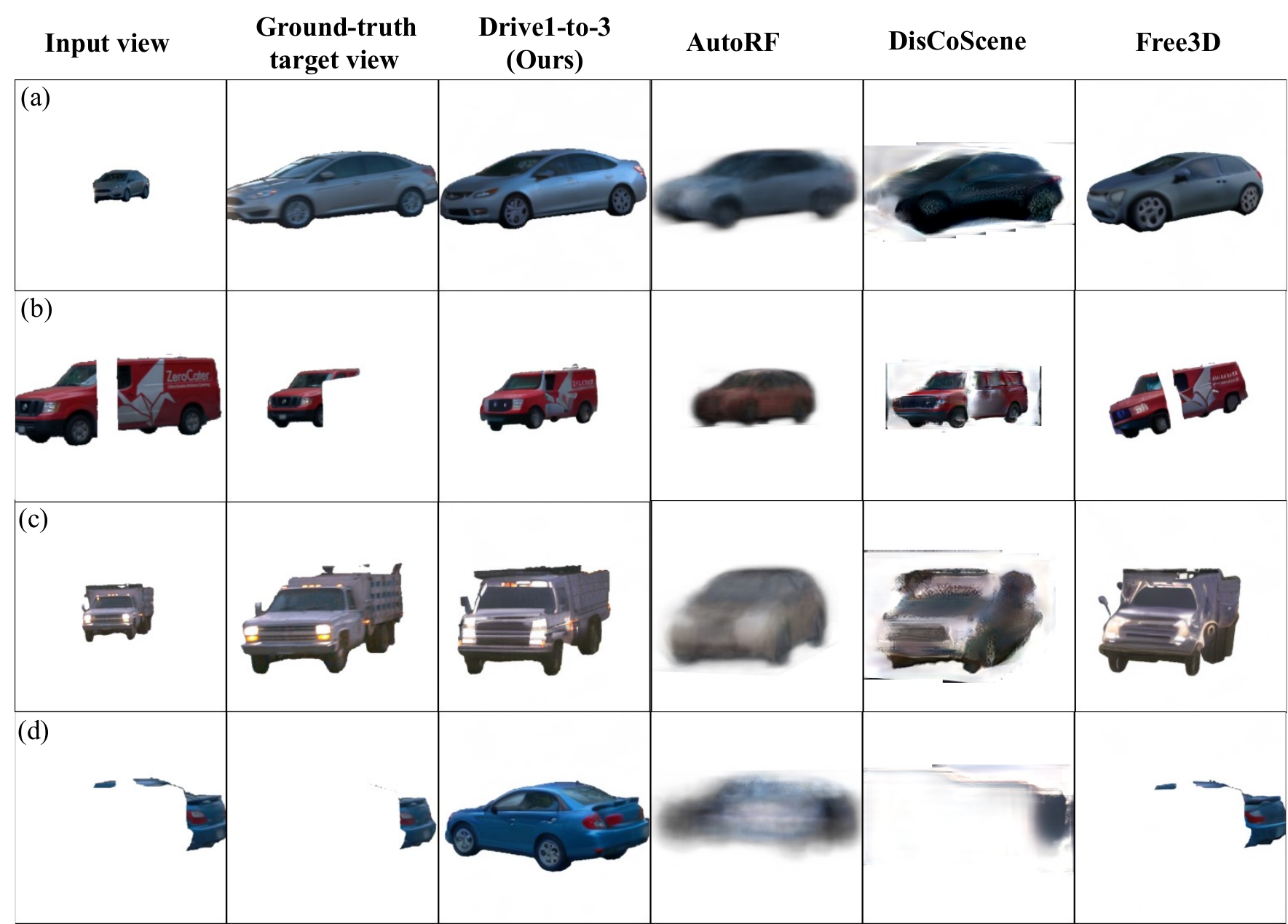}
  \centering
  \caption{Qualitative comparison of our method with AutoRF~\citep{muller2022autorf}, DisCoScene~\citep{xu2023discoscene} and Free3D~\citep{zheng2023free3d} on real vehicle images.}
  \label{fig:exp_compautorfdiscoscene}   
\end{figure}

\subsection{Experimental Details}
\label{sec:exp_details}

\textbf{Datasets.}  
We primarily utilize the Waymo Open Dataset~\citep{sun2020scalability} for evaluation, 
focusing specifically on its object asset dataset curated by \cite{shen2023gina}.
The training set consists of 18,842 vehicle objects, where we collect object images from the video frames with every 3 degree viewpoint change in azimuth, resulting in 131,987 training images.
The validation set comprises 4,808 vehicle objects, with one frame randomly selected as the source view and five additional frames as the target views, totaling 24,040 frames. We also test on the Nuscenes~\cite{caesar2020nuscenes} and DVM-Car~\cite{huang2022dvm} dataset for demonstrating cross-dataset generalization. 

\begin{table}[t]
  \centering
  \caption{Quantitative comparison with AutoRF and DisCoScene.}
  \label{tab:main_results}
  \resizebox{0.9\linewidth}{!}{
  \begin{tabular}{c|c|ccc|c}
    \toprule
    Method & Angles & PSNR$\uparrow$ & SSIM$\uparrow$ & LPIPS$\downarrow$ & FID$\downarrow$ \\
    \midrule    
    \multirow{4}*{{AutoRF~\cite{muller2022autorf}}} & All & \textbf{24.180} & 0.897 & 0.098 & \multirow{4}*{21.853}  \\[0.5ex]
    & 0-30 & 24.582 & 0.903 & 0.093  \\[0.5ex]
    & 30-60 & 21.454 & 0.857 & 0.130  \\[0.5ex]
    & 60-180 & 21.198 & 0.851 & 0.136  \\
    \midrule
    \multirow{4}*{{\shortstack{DisCoSence~\cite{xu2023discoscene}}}} & All & 22.230 & 0.886 & 0.091& \multirow{4}*{13.407}  \\[0.5ex]
    & 0-30 & 22.983 & 0.897 & 0.081  \\[0.5ex]
    & 30-60 & 17.217 & 0.812 & 0.151  \\[0.5ex]
    & 60-180 & 16.324 & 0.794 & 0.175  \\    
    \midrule    
    \multirow{4}*{{\shortstack{Ours: \\ Drive-1-to-3}}} & All & 23.632 & 0.896 & 0.064 & \multirow{4}*{4.181}\\[0.5ex]
    & 0-30 & 24.403 & 0.909 & 0.056  \\[0.5ex]
    & 30-60 & 18.421 & 0.810 & 0.119  \\[0.5ex]
    & 60-180 & 17.838 & 0.799 & 0.130 \\
    \midrule
    \multirow{4}*{{\shortstack{Ours: \\ Drive-1-to-3\\+LGM~\cite{tang2024lgm}}}} & All & 23.995 & \textbf{0.912} & \textbf{0.057} & \multirow{4}*{\textbf{4.122}}\\[0.5ex]
    & 0-30 & 24.664 & 0.922 & 0.050 \\[0.5ex]
    & 30-60 & 19.526 & 0.847 & 0.100 \\[0.5ex]
    & 60-180 & 18.798 & 0.834 & 0.116 \\    \bottomrule
  \end{tabular}}
\end{table}

\noindent  \textbf{Metrics.} 
We follow~\citep{zheng2023free3d,liu2023zero,deitke2024objaverse} to assess novel view synthesis quality at target views by PSNR, SSIM, and LPIPS~\citep{zhang2018unreasonable}. Note the ground truth image may present with occlusion whereas our method predicts the entire vehicle, causing a bias in evaluation; hence, we mask out occluded pixels in our output before comparing with the ground truth.  
We also render $360^\circ$ views at 90-degree intervals in azimuth, totaling 4 images, and evaluate using dataset-level Fréchet Inception Distance (FID)~\citep{heusel2017gans} in the absence of ground truth.

\vspace{1mm}
\noindent \textbf{Baselines.} 
We compare our work against: 1) the single-view 3D reconstruction method AutoRF~\cite{muller2022autorf}, which uses an encoder-decoder network to predict a NeRF representation for real vehicles; 2) the state-of-the-art GAN-based  method DisCoScene~\cite{xu2023discoscene} for real vehicles, where we perform GAN inversion~\cite{roich2022pivotal} to generate a NeRF from the input image.

\vspace{1.5mm}
\noindent \textbf{Combination with LGM for 3D reconstruction.} We render $360^\circ$ views from Drive-1-to-3 and feed them to LGM~\cite{tang2024lgm} for 3D reconstruction as a Gaussian Splatting, which in turn can be rendered from any viewpoints. We denote results from this step as `Drive-1-to-3+LGM'. For improved performance, we also finetune LGM on the Waymo dataset, as detailed in the supplementary material.

\begin{table}[t]
\centering
\caption{Ablation study of \Ours on the impact of rich priors from pretrained models,  occlusion-aware training (\textit{Occ}), symmetric prior (\textit{Sym}), relative versus absolute pose in the global (\textit{Glob-Pose}) and local (\textit{Loc-Pose}) pose condition.} 

  \label{tab:ablationnvs}
  \resizebox{\linewidth}{!}{
  \begin{tabular}{lccccc|c|ccc|c}
     \toprule
    & Pretrain & Occ. & Sym.  & Glob-Pose & Loc-Pose & Angles & PSNR$\uparrow$ & SSIM$\uparrow$ & LPIPS$\downarrow$  & FID$\downarrow$ \\
     \midrule
    \multirow{4}*{(a)} & & \multirow{4}*{\checkmark} & \multirow{4}*{\checkmark} & \textcolor{purple}{\multirow{4}*{Relative}} & \textcolor{purple}{\multirow{4}*{Relative}} & all & 15.542 & 0.763 & 0.206 & \multirow{4}*{11.053}\\[0.2ex]
    & & & & & & 0-30 & 15.816 & 0.772 & 0.199   \\[0.2ex]
    & & & & & & 30-60 & 13.780 & 0.703 & 0.247 \\[0.2ex]
    & & & & & & 60-180 &13.229 & 0.699 & 0.256 \\[0.2ex]
    \noalign{\vskip -2mm}
    \multicolumn{11}{c}{\begin{tikzpicture}
        \draw[dashed] (0,0) -- (16,0);
    \end{tikzpicture}} \\[0.2ex]
    \multirow{4}*{(b)} &  & \multirow{4}*{\checkmark} & \multirow{4}*{\checkmark} & \textcolor{cvprblue}{\multirow{4}*{Absolute}} & \textcolor{cvprblue}{\multirow{4}*{Absolute}}  & all & 17.356 & 0.803 & 0.157 & \multirow{4}*{12.965} \\[0.2ex]
    & & & & & & 0-30 & 17.633 & 0.811 & 0.151 \\[0.2ex]
    & & & & & & 30-60 & 15.467 & 0.746 & 0.194  \\[0.2ex]
    & & & & & & 60-180 & 15.339 & 0.744 & 0.197\\
    \midrule
    \multirow{4}*{(c)} & \multirow{4}*{\checkmark} & & & \textcolor{purple}{\multirow{4}*{Relative}} & \textcolor{purple}{\multirow{4}*{Relative}} & all & 21.390 & 0.841 & 0.119 & \multirow{4}*{8.871}  \\[0.2ex]
    & & & & & & 0-30 & 22.497 & 0.868 & 0.096 \\[0.2ex]
    & & & & & & 30-60 & 13.641 & 0.651 & 0.281  \\[0.2ex]
    & & & & & & 60-180 & 13.829 & 0.661 & 0.271  \\[0.2ex]
    \noalign{\vskip -2mm}
    \multicolumn{11}{c}{\begin{tikzpicture}
        \draw[dashed] (0,0) -- (16,0);
    \end{tikzpicture}} \\[0.2ex]
    \multirow{4}*{(d)} &\multirow{4}*{\checkmark} & \multirow{4}*{\checkmark} & & \textcolor{purple}{\multirow{4}*{Relative}} & \textcolor{purple}{\multirow{4}*{Relative}} & all & 22.599 & 0.867 & 0.093 & \multirow{4}*{7.090}\\[0.2ex]
    & & & & & & 0-30 & 23.523 & 0.885 & 0.079  \\[0.2ex]
    & & & & & & 30-60 & 16.272 & 0.740 & 0.190  \\[0.2ex]
    & & & & & & 60-180 & 15.909 & 0.731 & 0.198 \\
    \midrule
    \multirow{4}*{(e)} &\multirow{4}*{\checkmark} & \multirow{4}*{\checkmark} & \multirow{4}*{\checkmark} & \textcolor{cvprblue}{\multirow{4}*{Absolute}}  & \textcolor{cvprblue}{\multirow{4}*{Absolute}}   & all &20.644 & 0.841 & 0.113 & \multirow{4}*{5.608}\\[0.2ex]
    & & & & & & 0-30 & 21.239 & 0.854 & 0.105   \\[0.2ex]
    & & & & & & 30-60 & 16.804 & 0.762 & 0.166   \\[0.2ex]
    & & & & & & 60-180 & 15.599 & 0.731 & 0.201 \\[0.2ex]
    \noalign{\vskip -2mm}
    \multicolumn{11}{c}{\begin{tikzpicture}
        \draw[dashed] (0,0) -- (16,0);
    \end{tikzpicture}} \\[0.2ex]
    \multirow{4}*{(f)} &\multirow{4}*{\checkmark} & \multirow{4}*{\checkmark} & \multirow{4}*{\checkmark} & \textcolor{purple}{\multirow{4}*{Relative}} & \textcolor{cvprblue}{\multirow{4}*{Absolute}}  & all & 23.813 & 0.899 & 0.060 & \multirow{4}*{4.813} \\[0.2ex]
    & & & & & & 0-30 & 24.482 & 0.910 & 0.054 \\[0.2ex]
    &  & & & & & 30-60 & 19.343 & 0.826 & 0.101 \\[0.2ex]
    &  & & & & & 60-180 & 18.610 & 0.814 & 0.113 \\
    \midrule
    \multirow{4}*{(g)} &\multirow{4}*{\checkmark} & \multirow{4}*{\checkmark} & \multirow{4}*{\checkmark} & \textcolor{purple}{\multirow{4}*{Relative}} & \textcolor{purple}{\multirow{4}*{Relative}} & all & 23.632 & 0.896 & 0.064 & \multirow{4}*{4.181}\\[0.2ex]
    &  & & & & & 0-30 & 24.403 & 0.909 & 0.056  \\[0.2ex]
    &  & & & & & 30-60 & 18.421 & 0.810 & 0.119  \\[0.2ex]
    &  & & & & & 60-180 & 17.838 & 0.799 & 0.130 \\
    \bottomrule
  \end{tabular}
  }
\end{table}

\begin{figure*}[t]
\vspace{-4mm}
    \begin{minipage}{0.34\linewidth} 
    \includegraphics[width=1.0\linewidth, trim = 0mm 0mm 0mm 0mm, clip]{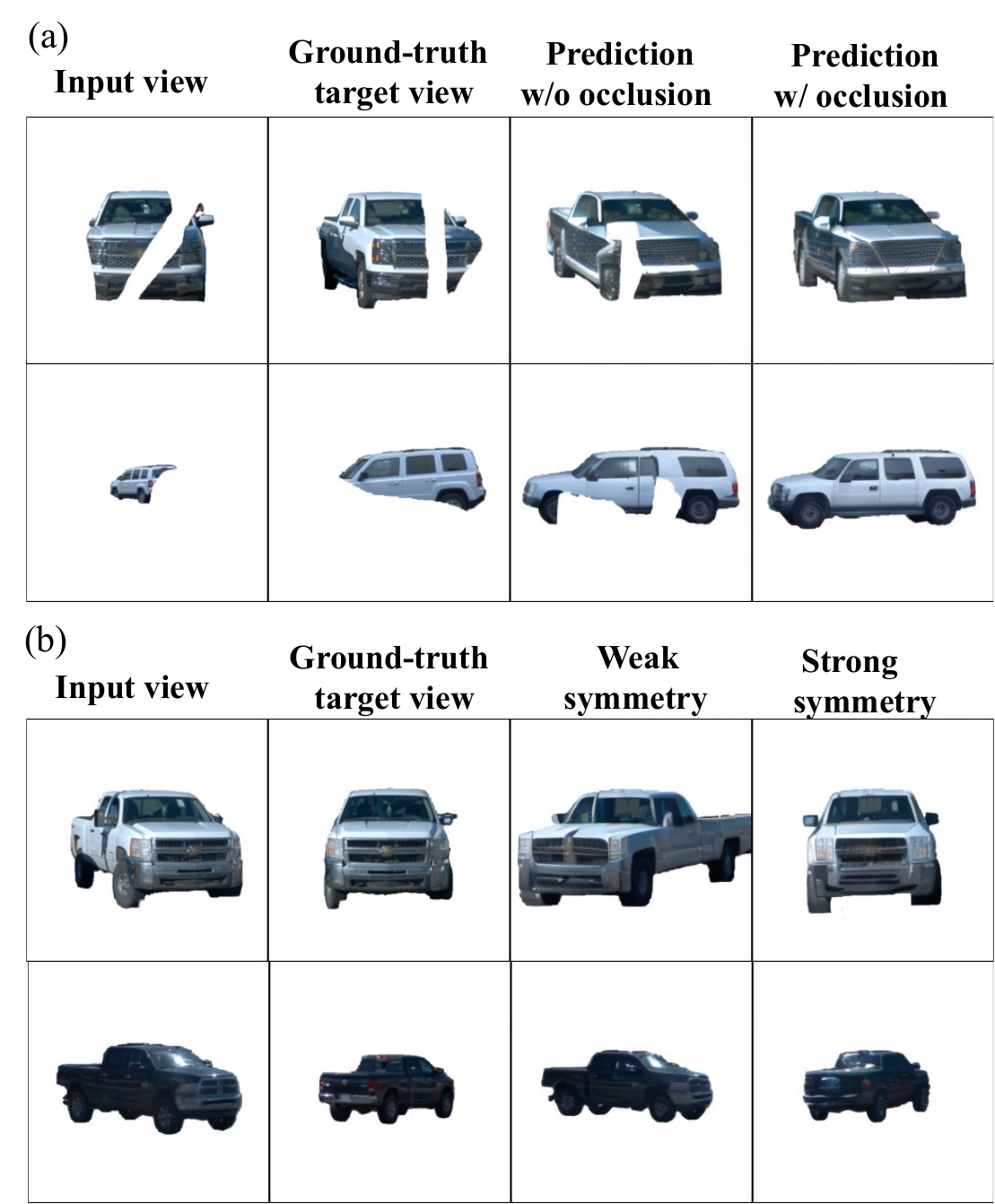}
      \centering
      \caption{Qualitative results showing benefits of (a) occlusion handling and (b) symmetric prior.}
    \label{fig:exp_occlusionsymmetry}   
   \end{minipage}\hfill 
   \begin{minipage}{0.64\linewidth} 
    \centering
    \includegraphics[width=1.0\linewidth, trim = 0mm 0mm 0mm 0mm, clip]{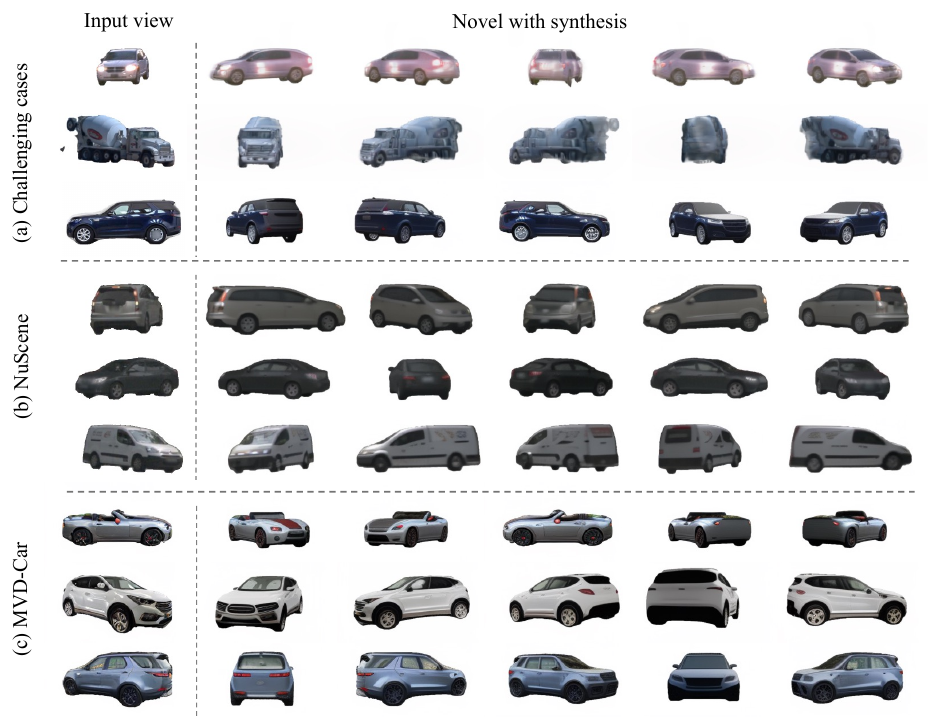}
    \centering
    \caption{Example results on (a) challenging cases, including night-time condition, unusual vehicles, and strong reflections; (b)(c) cross-dataset generalization on Nuscenes and MVD-Car dataset.}    
    \label{fig:moreevaluations}   
   \end{minipage}    
\vspace{-3mm}
\end{figure*}

\subsection{Comparisons on Novel View Synthesis}
\label{sec:comparison}
\textbf{Qualitative comparisons.} We start by qualitative comparison with AutoRF~\cite{muller2022autorf}, DisCoScene~\cite{xu2023discoscene} and Free3D~\cite{zheng2023free3d} to highlight our advantage in image quality. 
As shown in \cref{fig:exp_compautorfdiscoscene}, one observes that while AutoRF is capable of predicting the global shape and appearance of the object thanks to its learned categorical prior, it fails to capture fine detailed textures despite its latent code being test-time optimized on the input view. With explicit generative modeling towards photorealism, DisCoScene exhibits higher image quality, but still largely lags behind our method. This is especially true for challenging cases such as the long-tail red truck in \cref{fig:exp_compautorfdiscoscene} (b) and the presence of large occlusion in (d). This is attributed to the rich image and 3D prior that our method derives from the large pretrained diffusion models. We also show results from original Free3D model, which tend to be cartoonish with inaccurate poses, and does not handle any occlusions. More comparisons are provided in the supplementary material.

\noindent \textbf{Quantitative comparisons.}
We report the results in \cref{tab:main_results};
besides metrics on the entire validation set, we also group them by the degree of viewpoint change based on the relative azimuth. 
As shown, Drive-1-to-3 achieves a notably lower FID score on $360^\circ$ view rendering as well as the LPIPS w.r.t. the ground truth target views, even under large viewpoint changes. This validates the high visual fidelity of our synthesized views. We note that AutoRF achieves relatively high PSNR/SSIM as it uses ground truth 3D box (i.e. object pose and size) to position the NeRF, ensuring accurate pixel alignment without the need to learn pose conditioning. However, its significantly higher LPIPS and FID indicate poor perceptual quality. Despite the need to learn pose consistency, Drive-1-to-3 achieves comparable PSNR/SSIM while retaining high image quality. Further, combining Drive-1-to-3 with LGM further improve performance. This demonstrates the utility of Drive-1-to-3 in providing consistent multiview input to the downstream 3D reconstruction model; we will discuss this further in \cref{sec:lgm}.

\subsection{Ablation Study}
\label{ab_study}
In \cref{tab:ablationnvs}, we perform ablation study for \Ours under various configurations to understand its behavior. 

\vspace{1.5mm}
\noindent \textbf{Impact of pre-trained models.} Instead of finetuning pretrained pose-conditioned diffusion models, we train \Ours from the original Stable Diffusion to learn the pose conditioning in real data from scratch. This way, we may also apply absolute pose conditioning as opposed to the relative one in pretrained models, without concern for sacrificing priors. As shown in \cref{tab:ablationnvs}(a)(b), their results are considerably inferior to the one with pretained model in (g). This underscores the merit to reap benefits from the rich diffusion prior for our domain-specific application. 

\vspace{1.5mm}
\noindent \textbf{Impact of occlusion handling.} We first demonstrate its impact via qualitative results in \cref{fig:exp_occlusionsymmetry}(a); more examples are in supplementary. Without occlusion-aware training, one observes the evident artifacts of translating the white occluded regions from the input to the target view, which are removed by the occlusion handling. Comparing \cref{tab:ablationnvs}(c) and (d), one observes that the occlusion handling improves the results quantitatively as well.

\vspace{1.5mm}
\noindent \textbf{Impact of symmetric prior.} 
Comparing \cref{tab:ablationnvs}(d) and (g), we observe quantitative improvements provided by the symmetric prior, especially under large viewpoint changes. In addition, we demonstrate the effectiveness of the strong symmetric guidance (cf. \cref{sec:symmetry}) against the weak one in \cref{fig:exp_occlusionsymmetry}(b) -- the strong guidance leads to superior pose conditioning accuracy. More comparisons are in supplementary.    

\vspace{1.5mm}
\noindent \textbf{Relative vs. absolute pose conditioning.} As discussed in \cref{sec:method_pose}, we may change the relative pose conditioning to an  absolute pose layer, while still loading other pretrained UNet parameters. This is applicable to both the global pose condition in Zero-1-to-3 and the local one added by Free3D, as depicted in \cref{sec:overview}. Comparing \cref{tab:ablationnvs} (f) to (g) reveals that this modification in the local pose condition results in improvements in PSNR/SSIM/LPIPS, but drops in FID for $360^\circ$ novel views. Further applying this to the global pose condition in (e) leads to degradation in all metrics. Overall, this highlights that the rich priors in pretrained models outweigh the benefits of absolute pose information.

\begin{figure*}[t]
  \centering
  \vspace{-2mm}
  \includegraphics[width=0.85\linewidth, trim = 0mm 0mm 0mm 0mm, clip]{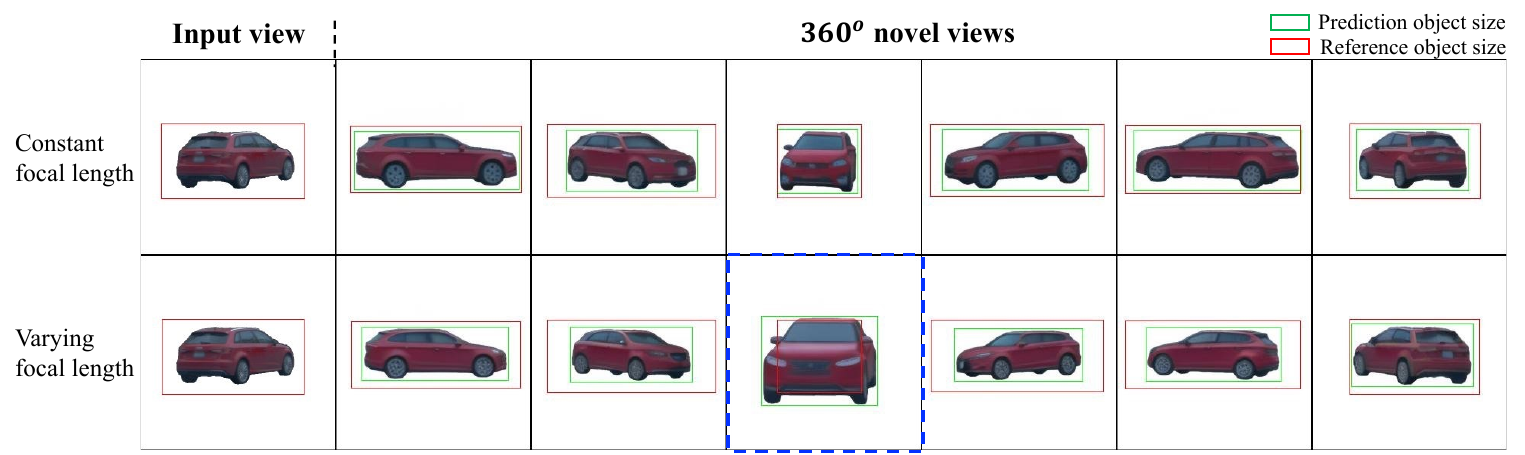}
  \centering
\vspace{-2mm}
  \caption{Qualitative comparison of varying object scales and varying focal lengths. The example in the blue dashed box shows a significant discrepancy between the prediction object size and the reference object size. Please see the supplementary for more rendering videos.}
  \label{fig:exp_sizeconsistency}   
\end{figure*}

\begin{figure*}[t]
   \begin{minipage}{0.64\linewidth} 
    \centering
    \includegraphics[width=1.0\linewidth, trim = 0mm 0mm 0mm 0mm, clip]{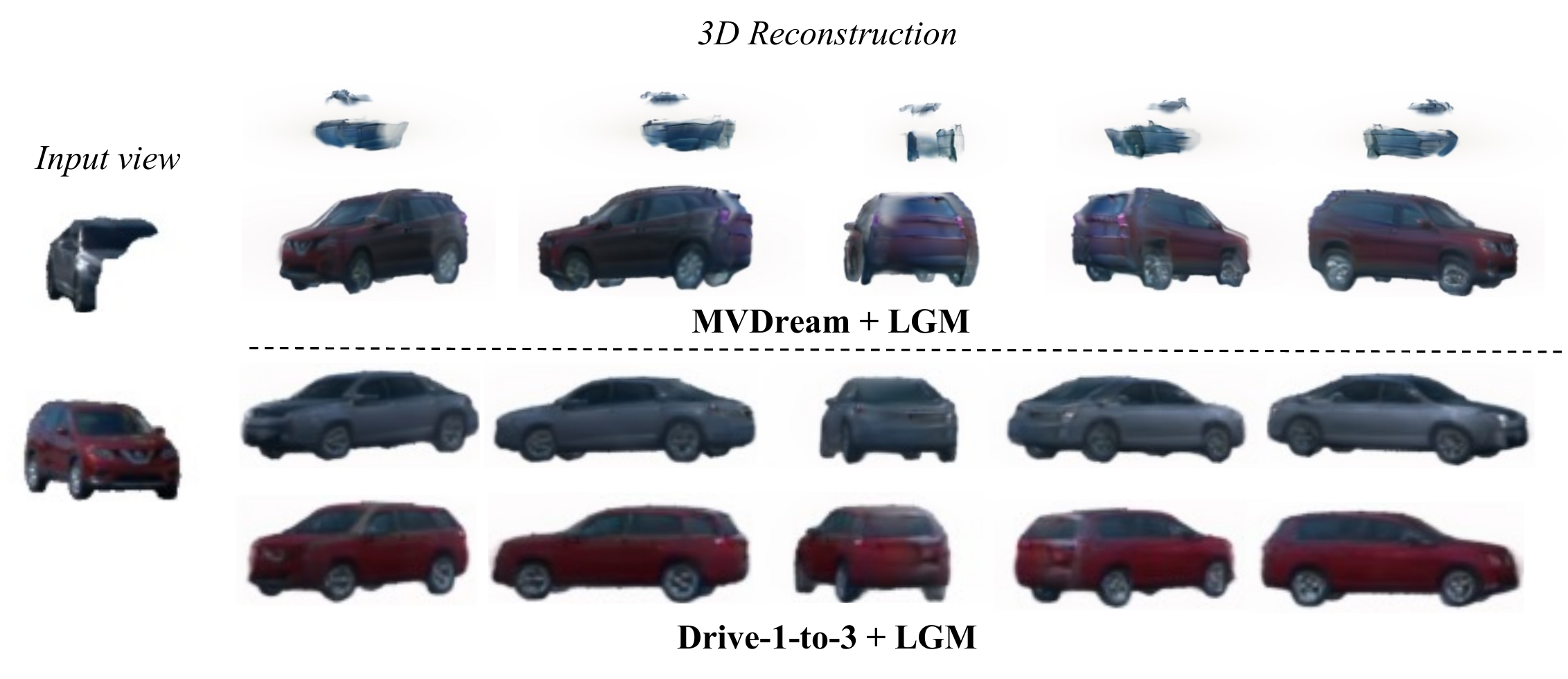}
    \centering
    \caption{LGM 3D reconstruction comparison with Drive-1-to-3 versus its default MVDream~\cite{shi2023mvdream} as the multiview generator. }    
    \label{fig:exp_lgm}   
   \end{minipage}\hfill
   \begin{minipage}{0.34\linewidth} 
    \includegraphics[width=1.0\linewidth, trim = 0mm 0mm 0mm 0mm, clip]{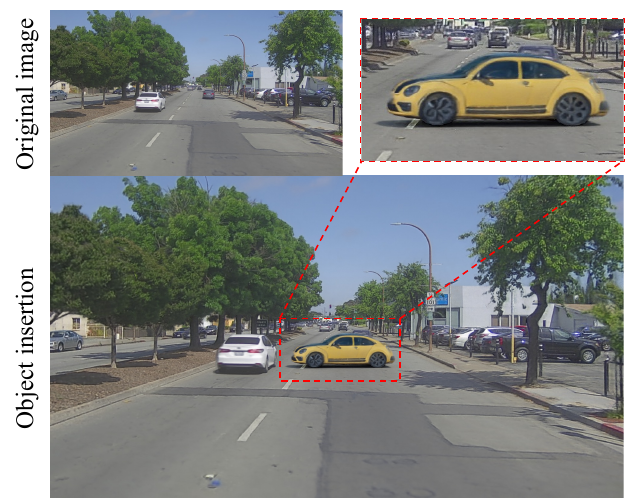}
  \centering
\vspace{-1mm}
  \caption{Application of Drive-1-to-3 for virtual object insertion in driving scenes.}
    \label{fig:objectinsertion}   
   \end{minipage}    
\end{figure*}

\vspace{1.5mm}
\noindent \textbf{Varying object scales vs. varying focal lengths.} 
Here we investigate the two cropping strategies discussed in \cref{sec:method_pose}.
Empirically, we find the diffusion model robust in learning across varying object scales but often struggles in maintaining object size consistency when training with varying focal lengths. We demonstrate this phenomenon with examples in \cref{fig:exp_sizeconsistency} utilizing the ``ground truth" object size -- we project the object 3D bounding boxes into the novel views to obtain their 2D boxes, which serve as references for the enclosing 2D boxes of the synthesized objects. As highlighted, one observes the large box size deviations in some views under varying focal lengths, while robust size consistency under constant focal length, despite its varying object scales in training data. We provide rendering videos in the supplementary to better illustrate this phenomenon. We attribute this to the strong image representation learned across scales in Stable Diffusion, yet the pose-conditioned diffusion model lacks learning across varying focal lengths.

\subsection{More Evaluations}

\noindent \textbf{Challenging cases.} In \cref{fig:moreevaluations}(a), we show results on hard cases where failure tends to arise, including night-time condition, unusual vehicle, and strong surface reflection. While the performance degrades as expected, Drive-1-to-3 still captures the object structure in a photorealistic manner.

\noindent \textbf{Cross-dataset generalization} In \cref{fig:moreevaluations}(b)(c), we demonstrate results on  the low-resolution NuScenes~\cite{caesar2020nuscenes} images and high-resolution DVM-Car~\cite{huang2022dvm} images, both yielding high-quality view synthesis results. More evaluations are given in the supplementary material.

\subsection{3D Reconstruction with LGM}
\label{sec:lgm}
As described in \cref{sec:exp_details}, our novel synthesis may serve as the input to LGM for 3D reconstruction. 
In \cref{fig:exp_lgm}, we compare with its default multiview generator MVDream~\cite{shi2023mvdream} and demonstrate superior reconstruction quality, underscoring the value of our approach. Note that we apply the same Waymo-finetuned LGM model to both for fair comparison.

\subsection{Application for Object Insertion}
Here, we demonstrate the application in virtual object insertion, by integrating the 3D object model from Drive-1-to-3+LGM output into Unisim~\cite{yang2023unisim}, a NeRF simulator for dynamic driving scenes.
We show an example result in \cref{fig:objectinsertion} to illustrate photorealistic insertion; we also provide the full video in the supplementary. Note that we also render object shadow by accounting for environment lighting. Please see the supplementary for more technical details.

%% file: sec/5_conclusion.tex
\section{Conclusion}

In this work, we study good practices to leverage rich priors in large pretrained diffusion models towards novel view synthesis for real vehicles. Our proposed framework handles the discrepancy between real and synthetic data from various aspects, including camera poses, object distances, occlusions, and viewpoint changes. It enables high-quality novel view synthesis without requiring large computing resource. A limitation of our model lies in lacking physical understanding of the object appearance such as albedo, material property, and light transport. This remains an important future direction as it facilitates the integration with graphics rendering pipelines for more complex simulation.

%% file: sec/X_suppl.tex
\clearpage
\newpage
       \twocolumn[
        \centering
        \Large
        \textbf{Supplementary Material}\\        
        \vspace{1.0em}
       ]

\section{Supportive Explanations}
\begin{figure}[t]
  \centering    \includegraphics[width=1.0\linewidth, trim = 0mm 0mm 0mm 0mm, clip]{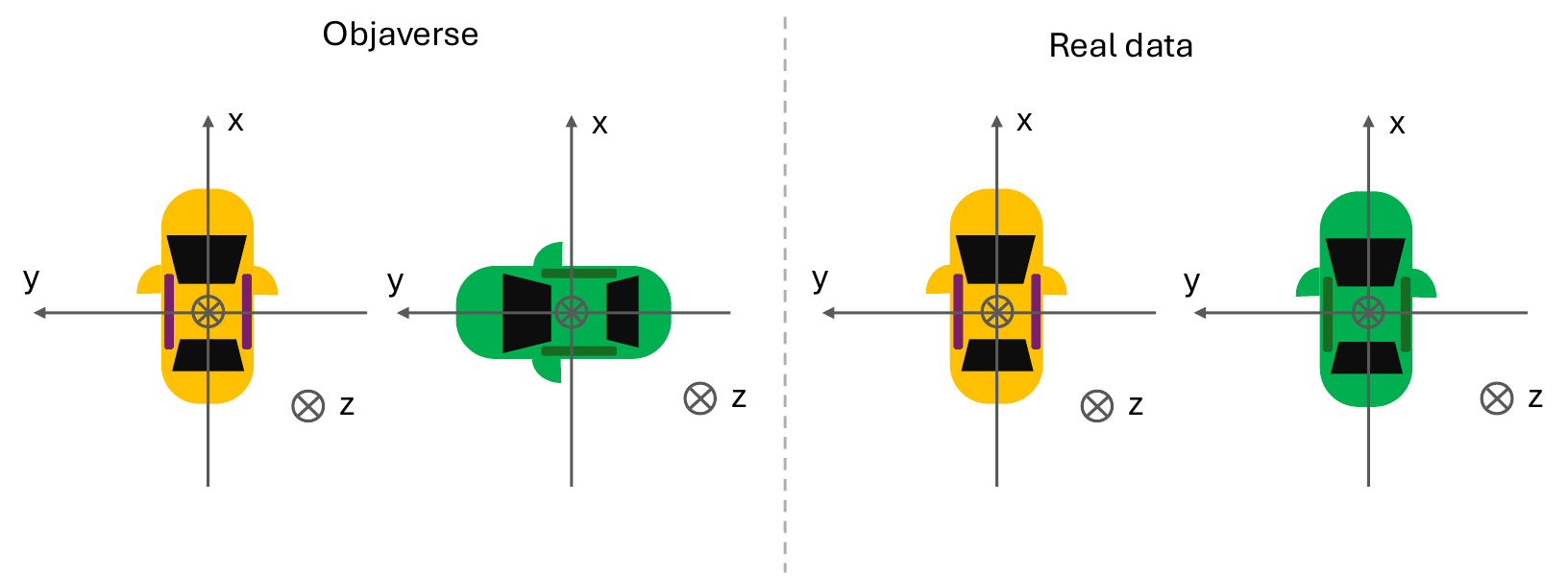}
  \centering
  \caption{Illustration of unaligned azimuth in Objaverse vs. the aligned azimuth in real data.}
  \label{fig:supp_relativeabsolute}   
\end{figure}

\noindent \textbf{More explanations on relative pose vs. absolute pose. }The reason that this matter arises lies in that, the azimuth angles 
 of the Objaverse data are not defined w.r.t a common reference, as illustrated in \cref{fig:supp_relativeabsolute}. Specifically, the origin of the object coordinate system is always set at the object center, and its z-axis always points upwards along the negative gravity direction, yielding aligned elevation definition across different objects. However, the azimuth definition along the horizontal direction is not consistent across objects—this is the crux of the problem. For example, for two different models in the car category, the x-axis may point to the front view of one car whereas to the side view of the other. Hence, taking the absolute poses for the source and target views as input may confuse the network training. But it is not a problem for the relative pose 
 as it is independent of the reference coordinate. As a result, existing pose-conditioned diffusion models (e.g. Zero-1-2-3) are mostly trained with the relative pose instead of the absolute pose.

\begin{figure}[t]
  \centering    \includegraphics[width=0.9\linewidth, trim = 0mm 0mm 0mm 0mm, clip]{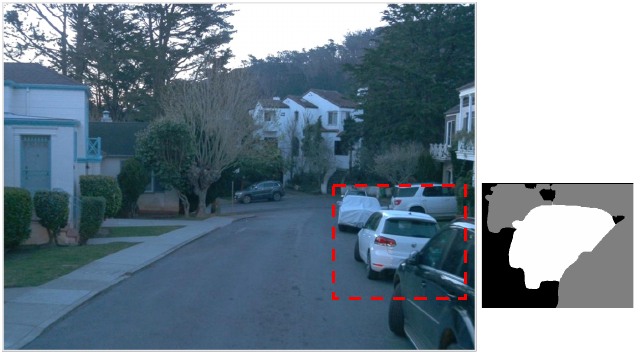}
  \centering
  \caption{Example occlusion mask, where white pixels indicate the foreground object, black pixels indicate background, and gray pixels indicate unknown regions where occlusion may arise.}
  \label{fig:supp_occlusionmask}   
\end{figure}

\begin{figure}
  \centering    \includegraphics[width=1.0\linewidth, trim = 0mm 0mm 0mm 0mm, clip]{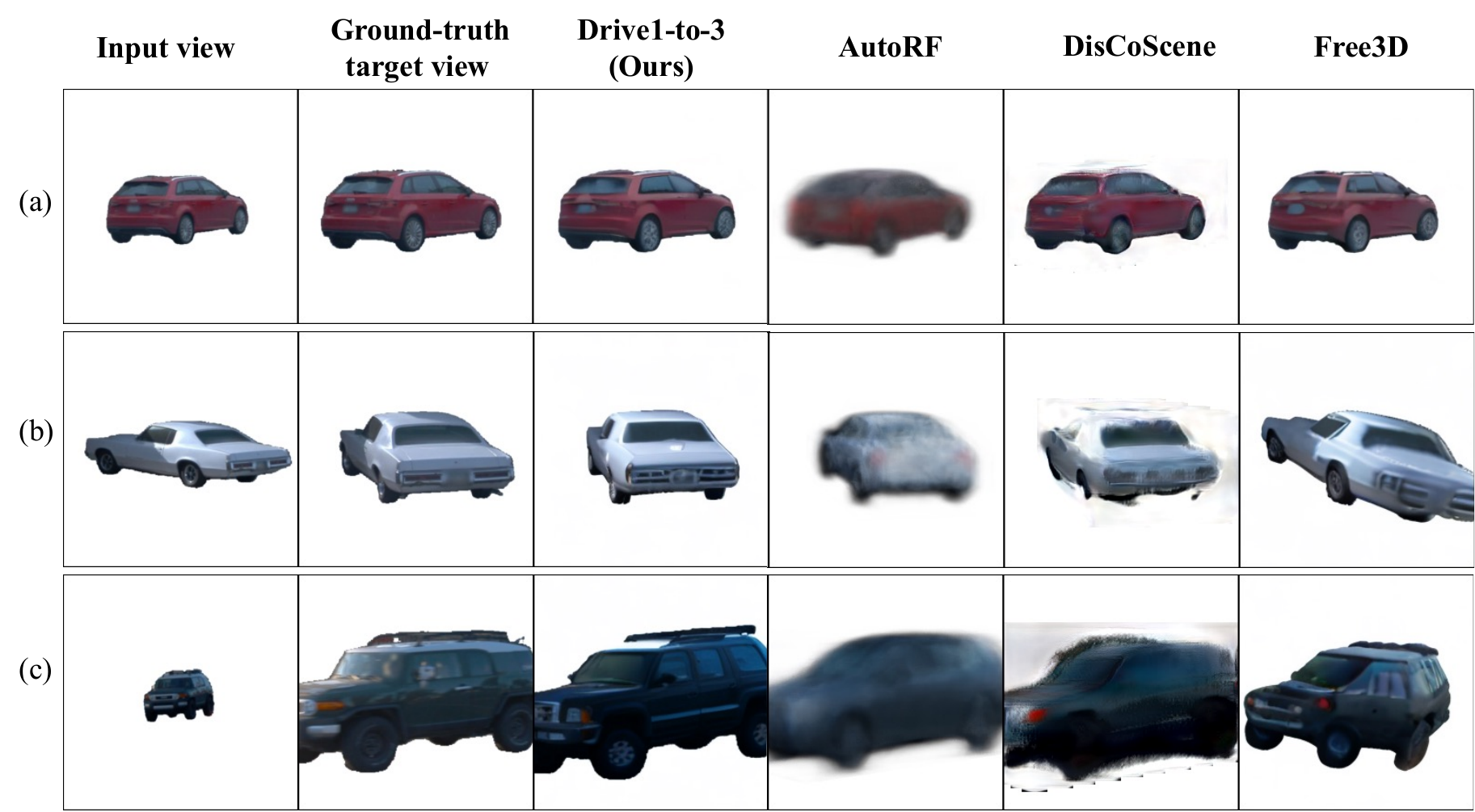}
  \centering
  \caption{Additional qualitative comparison of our method with AutoRF~\cite{muller2022autorf},
DisCoScene~\cite{xu2023discoscene} and Free3D~\cite{zheng2023free3d} on real vehicle images.}
  \label{fig:supp_qualitative}   
\end{figure}

\begin{figure*}[t]
  \centering    \includegraphics[width=0.85\linewidth, trim = 0mm 0mm 0mm 0mm, clip]{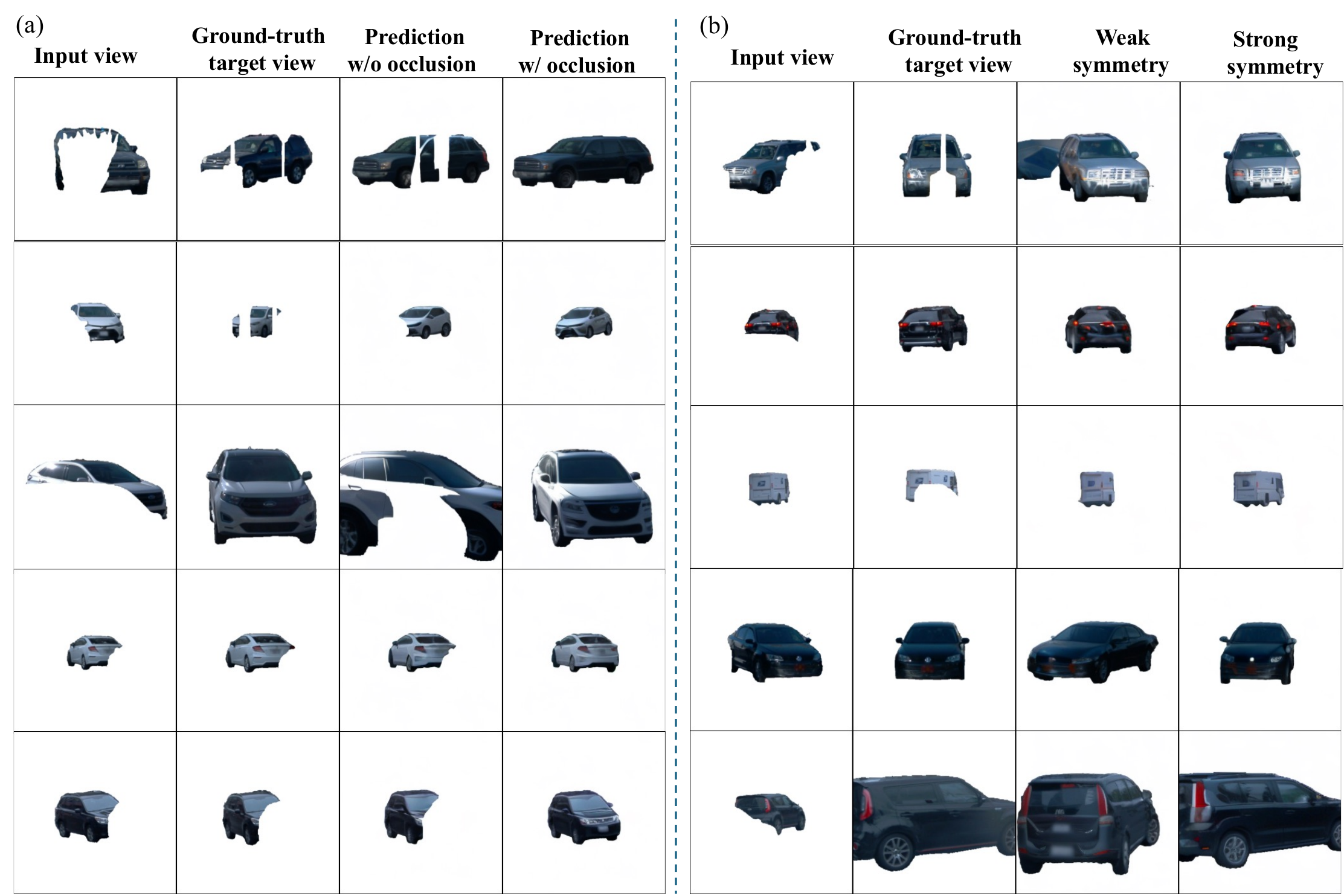}
  \centering
  \caption{Additional results showing benefits
of (a) occlusion handling and (b) symmetric prior.}
  \label{fig:supp_occlusionsymmetry}   
\end{figure*}

However, such ambiguity in pose definition does not exist for the real Waymo dataset. In contrast to the Internet-sourced 3D models in Objaverse, the object poses for all the vehicles in Waymo are human-annotated according to a common protocol. This leads to consistent pose definition, e.g. x-axis always points to the front view, y-axis to the left, and z-axis to the up. As such, it enables network training with the absolute pose as input, which is more informative than solely their relative difference. However, this also deviates from the model input used in the large pretrained model, hence may compromise the utility of its encoded prior in the new training setup. This gives the trade-off that we discussed in the main paper.

\vspace{0.2cm}
\noindent \textbf{Occlusion mask details.} We use the panoptic semantic masks provided by the Waymo dataset for extracting plausible occlusion masks. We define the following categories as potential occluding objects or things: \{\textit{car, truck, bus, other large vehicle, bicycle, motorcycle, trailer, pedestrian, cyclist, motorcyclist, bird, ground animal, construction cone pole, pole, pedestrian object, sign, traffic light, vegetation}\}.
We then generate the occlusion mask composed of surrounding pixels that fall into these categories. \cref{fig:supp_occlusionmask} shows an example, where the gray pixels correspond to regions where occlusion may arise.
Our occlusion-aware training is designed to exclude these potential occluding pixels in training loss. Note that our occlusion reasoning is conservative, as it may also exclude pixels that fall behind the foreground object in 3D space, but we found this strategy sufficient in our application.

\begin{figure*}[t]
  \centering    \includegraphics[width=0.85\linewidth, trim = 0mm 0mm 0mm 0mm, clip]{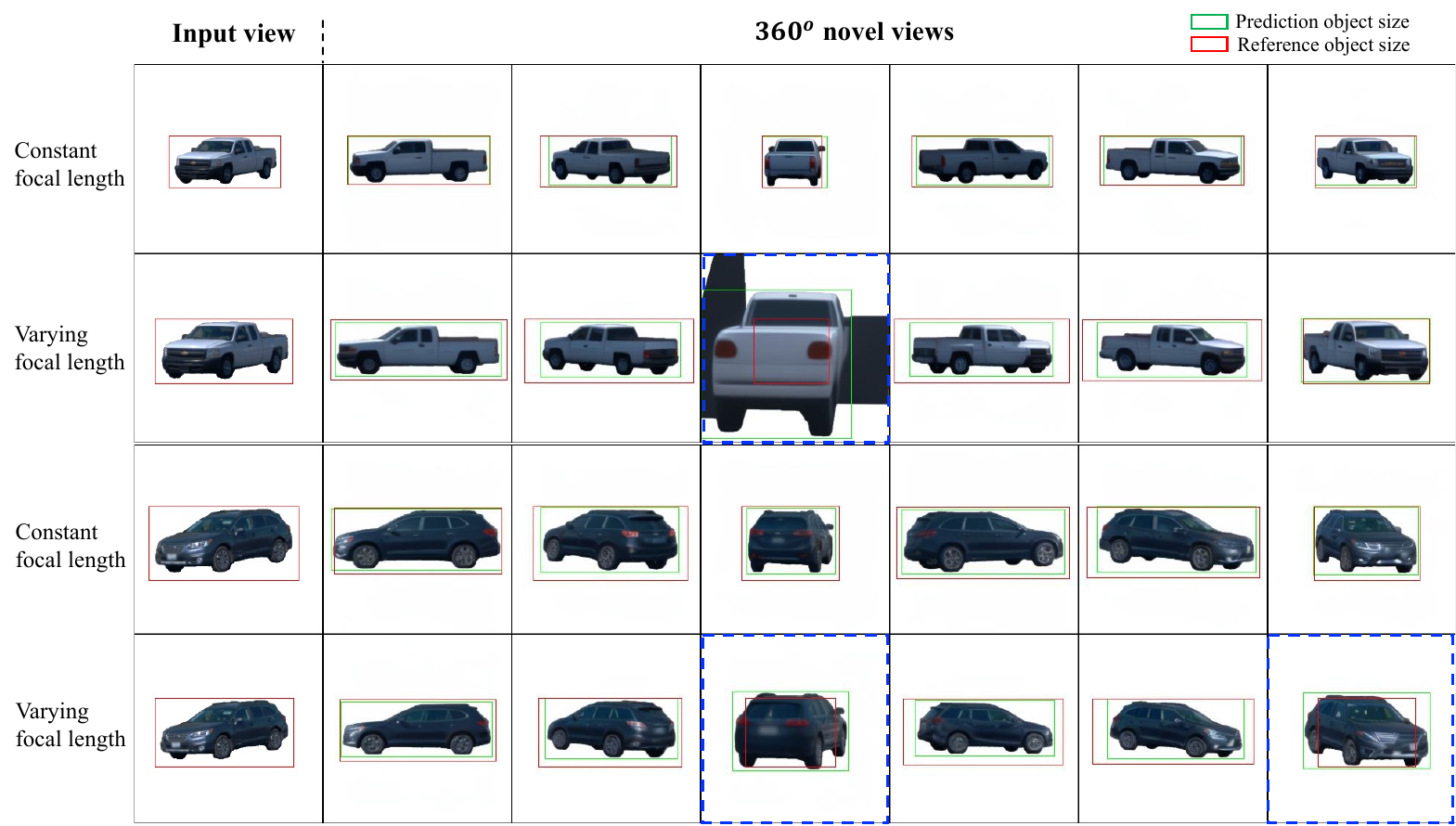}
  \centering
  \caption{Qualitative comparison of varying object scales and varying focal lengths. The examples in the blue dashed boxes show a significant
discrepancy between the prediction object size and the reference object size.}
  \label{fig:supp_objectsize}   
\end{figure*}

\begin{figure*}[t]
  \centering    \includegraphics[width=0.9\linewidth, trim = 0mm 0mm 0mm 0mm, clip]{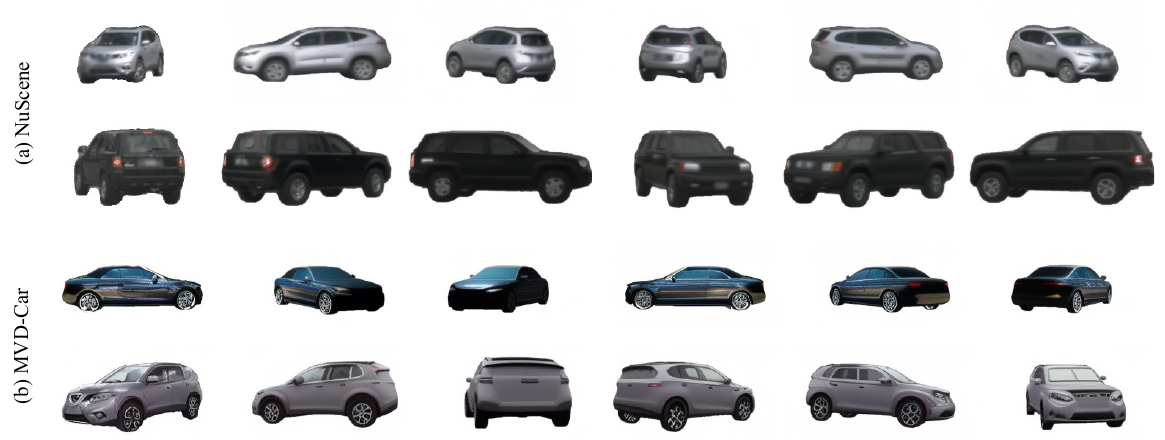}
  \centering
  \caption{Additional results on NuScenes and DVM-Car dataset.}
  \label{fig:supp_nuscenesdvm}   
\end{figure*}

\begin{table}[t]
  \centering
  \caption{Ablation Study on Multi-view Attention.}
  \label{tab:ab_multi-view}
  \resizebox{0.9\linewidth}{!}{
  \begin{tabular}{c|c|ccc|c}
    \toprule
    Method & Angles & PSNR$\uparrow$ & SSIM$\uparrow$ & LPIPS$\downarrow$ & FID$\downarrow$  \\
    \midrule    
    \multirow{4}*{{\shortstack{Ours}}} & All & \textbf{23.632} & \textbf{0.896} & \textbf{0.064} & \multirow{4}*{4.181}\\[0.5ex]
    & 0-30 & 24.403 & 0.909 & 0.056  \\[0.5ex]
    & 30-60 & 18.421 & 0.810 & 0.119  \\[0.5ex]
    & 60-180 & 17.838 & 0.799 & 0.130 \\
    \midrule 
    \multirow{4}*{{\shortstack{+ Multi-view Attention}}} & All & 23.408 & \textbf{0.896} & \textbf{0.064} & \multirow{4}*{\textbf{3.790}}\\[0.5ex]
    & 0-30 & 24.320 & 0.911 & 0.052  \\[0.5ex]
    & 30-60 & 17.346 & 0.793 & 0.142  \\[0.5ex]
    & 60-180 & 16.263 & 0.779 & 0.161 \\
    \bottomrule
  \end{tabular}}
  \label{tab:attention}  
\end{table}

\vspace{0.2cm}
\noindent \textbf{Finetune LGM.} As mentioned in Sec.~4.1 of the main paper, we finetune LGM on the Waymo object dataset. Similar to our Drive-1-to-3, we randomly select four different views of the object as the input, and another four views from the same sequence as the ground truth target views for training. After training, it takes four views generated from Drive-1-to-3 for reconstructing a Gaussian Splatting (GS) model of the vehicle; example results are demonstrated in the attached qualitativeresult.mp4.

\begin{table}[t]
  \centering
  \setlength{\tabcolsep}{18pt} 
  \caption{Quantitative comparison on  generalizing to DVM-Car Dataset.}
  \resizebox{0.5\linewidth}{!}{
    \begin{tabular}{cc}
    \toprule
    Method & FID$\downarrow$  \\
    \midrule 
    Free3D & 24.267\\
    Ours & 21.671\\
    \bottomrule
    \end{tabular}}
    \label{tab:dvmfid}
\end{table}

\vspace{0.2cm}
\noindent \textbf{Virtual object insertion.} We adopt Unisim~\cite{yang2023unisim} as the NeRF simulator for demonstrating object insertion. We run Drive-1-to-3+LGM to obtain the Gaussian Splatting model of the vehicle. To be compatible with the NeRF simulator, we render $360^\circ$ views from the GS model and train a NeRF for the vehicle, which can then be inserted into the scene following Unisim. To obtain shadow, we use LGM's code to convert the GS model to a mesh, then apply Blender to render the shadow on the ground plane, with a manually estimated lighting direction of the sun.

\vspace{0.2cm}
\noindent \textbf{Multi-view cross attention.} In our implementation, the multi-view cross attention proposed by Free3D is not used, as we did not observe noticeable improvements, as reported in \cref{tab:attention}.

\section{Additional Results}

\noindent \textbf{Additional qualitative comparisons.}
We provide additional qualitative comparisons with AutoRF, DisCoScene, and Free3D in \cref{fig:supp_qualitative}, with three examples to demonstrate the effectiveness of our method. Example (a) shows that all methods can handle a simple pose change when the car is a common type. In (b),
we see that when presented with a long-tailed vintage car, Free3D predicts the correct shape but the wrong pose, indicating that the diffusion model can provide a shape prior. Our method correctly predicts both, demonstrating that we leverage the diffusion prior to achieve novel view synthesis on real cars. Example (c) illustrates a case with a significant angle and distance change from the source to the target view. Again, AutoRF and DisCoScene results are very blurry, and Free3D, which was trained on Objaverse, produces cartoonish image with inaccurate pose. We also provide video comparison with Free3D in the attached qualitativeresult.mp4.

\vspace{0.2cm}
\noindent \textbf{Additional results on occlusion handling and symmetric.} In \cref{fig:supp_occlusionsymmetry}, panel (a) demonstrates that predictions with occlusion-aware training can more accurately reconstruct occluded parts of the cars. Panel (b) shows that incorporating a strong symmetry prior results in more precise predictions, particularly for significant pose changes.

\vspace{0.2cm}
\noindent \textbf{Additional examples with video for size inconsistency.}
We show additional examples in \cref{fig:supp_objectsize} that demonstrate the advantage of training with a constant focal length in object size consistency, in comparison with training with varying focal lengths. We also attach videos in qualitativeresult.mp4 for better illustration.

\vspace{0.2cm}
\noindent \textbf{Additional results on NuScenes and DVM-Car dataset} are shown in are shown in \cref{fig:supp_nuscenesdvm}. It demonstrates the generalization capability of Drive-1-to-3 to both low-resolution and high-resolution images, thanks to the rich diffusion prior it has leveraged. In \cref{tab:dvmfid}, we report the FID score on the DVM-Car dataset, in comparison with that from Free3D.

\vspace{0.2cm}
\noindent \textbf{Video for virtual object insertion.}
In the attached objectinsertion.mp4, we demonstrate virtual object insertion in real driving scenes to create safety-critical scenarios. We insert the vehicle either as a static one in the middle of the road or a dynamic one driving in the wrong direction. We demonstrate close-loop simulation by having the ego vehicle change lane to prevent collision.